\documentclass[journal,comsoc]{IEEEtran}
\usepackage[T1]{fontenc}
\usepackage{cite}

\ifCLASSINFOpdf
   \usepackage[pdftex]{graphicx}
\else
   \usepackage[dvips]{graphicx}
\fi

\usepackage{amsmath}
\usepackage[cmintegrals]{newtxmath}

\usepackage[numbers]{natbib}
\usepackage{booktabs}
\usepackage{multirow}
\usepackage{multicol}
\usepackage{amsmath}
\usepackage{algpseudocode}
\usepackage{mathtools}
\usepackage[linesnumbered,ruled,vlined]{algorithm2e}

\usepackage{tikz}

\mathchardef\mhyphen="2D

\begin{document}
%
\title{SCALE: Small-Model-First Concept Abstraction with LLM-Guided Experience Internalization}

\author{Hong~Su
\IEEEcompsocitemizethanks{\IEEEcompsocthanksitem H. Su is with the School of Computer Science, Chengdu University of Information Technology, Chengdu, China.\\
 E-mail: suguest@126.com. \\
\protect\\
}
\thanks{}}

\markboth{Journal of \LaTeX\ Class Files,~Vol.~14, No.~8, August~2015}%
{Shell \MakeLowercase{\textit{et al.}}: Bare Demo of IEEEtran.cls for IEEE Communications Society Journals}
%

\maketitle

\begin{abstract}
Autonomous agents must continually acquire reusable knowledge from experience without relying on costly large language model (LLM) reasoning for every decision. However, existing experience-based approaches often struggle to transform accumulated observations and reasoning into compact knowledge that lightweight models can independently apply. This paper proposes SCALE, a small-model-first framework that integrates selective LLM reasoning, experience-driven concept abstraction, and knowledge internalization. A lightweight model handles routine predictions, while prediction errors and accumulated cross-episode evidence selectively trigger LLM reflection. The LLM analyzes relevant historical experiences, generates positive and negative counterfactual examples, and proposes reusable concept hypotheses. Independently validated concepts are progressively incorporated into hierarchical neural representations, enabling lightweight models to exploit increasingly abstract knowledge. Meanwhile, mastered experiences are excluded from active memory while their original records remain available in a complete archive. Experiments across ten random seeds show that historical recall improves held-out accuracy from 87.11\% to 98.28\%, hierarchical concept knowledge increases difficult-case accuracy from 55.33\% to 66.47\%, and abstract archiving reduces active experience records from 6,000 to 1,528.
\end{abstract}


\IEEEpeerreviewmaketitle

\section{Introduction}
\label{sec:introduction}

Autonomous agents operating over long periods must make decisions from
changing observations while accumulating knowledge that can improve future
actions. Large language models (LLMs) provide flexible reasoning over
individual situations and can synthesize lessons from experience \cite{naveed2025comprehensive} \cite{zhao2026survey}, whereas
lightweight models offer a practical means of handling frequent decisions
without repeated expensive inference. The central challenge is to combine
these complementary capabilities so that experience produces \emph{lasting
improvements in the agent's own predictive knowledge}, rather than merely
providing more information for the next LLM request. This is especially
relevant when important evidence is separated by long intervals, new object
types share previously unseen conditions, and a continuing agent must retain
useful history without allowing its active memory to grow without bound.
Recent studies on efficient model selection, reflective agents, and
concept-guided prediction establish promising foundations for this goal
\cite{ong2025routellm,shinn2023reflexion}.

However, these research directions do not by themselves provide a complete
experience-to-knowledge pathway for a small-model-first agent. Cost-aware
routing chooses which model should answer a query, but does not necessarily
make the lightweight model more capable after the expensive intervention
\cite{xin2025hybrid,ong2025routellm}. Reflection and experience-memory
methods can retrieve useful past events or textual insights, yet these
insights often remain dependent on future LLM calls instead of becoming
lightweight predictive parameters
\cite{zhao2024expel,park2023generative}.
Conversely, concept-based predictors can make abstract representations
operational, but typically assume an available concept vocabulary or concept
supervision rather than deriving candidate concepts from the agent's own
temporally dispersed experiences
\cite{koh2020cbm,zarlenga2022cem}. Continual replay and organized external
memory help preserve past information, but retaining or retrieving examples
is not equivalent to discovering a shared condition that generalizes across
situations \cite{lopezpaz2017gem,xu2025amem}. Moreover, an LLM-generated
explanation or counterfactual cannot be assumed correct without independent
checks. These limitations motivate a framework in which selective reasoning,
validated abstraction, small-model learning, and long-term experience
management form one continuing process.

In this paper, we propose \emph{SCALE} (\textbf{S}mall-model-first
\textbf{C}oncept \textbf{A}bstraction with \textbf{L}LM-guided
\textbf{E}xperience Internalization). Its ordinary decision pathway uses a
lightweight task predictor, optionally assisted by concepts that the agent
has already learned. The LLM is not required for every observation: an
incorrect prediction identified through independent feedback may initiate
reasoning about its cause, while accumulated evidence spanning different
historical episodes may separately initiate a concept review. The agent
retrieves relevant verified experiences, and the LLM proposes an observable
explanation and positive/negative counterfactual learning examples. These
proposals are admitted for training only after checking their consistency
with independent task and event semantics. Thus, SCALE uses expensive
reasoning to create reusable training knowledge rather than substituting an
LLM answer for every small-model decision.

Beyond individual error-driven learning, SCALE uses retrospective evidence
to form hierarchical concepts that can guide later predictions. A candidate
concept describes common required and excluded observable conditions, is
tested against independent positive and negative cases, and becomes a
trainable neural parent only after validation. For example, risk experiences
involving sharp glass and stone fragments may support an initial shared
concept; later encounters with flexible sharp fragments may motivate a
broader hypothesis that does not require rigidity. Such a hypothesis is not
accepted simply because the LLM proposes it, nor is a higher abstraction
assumed to be more accurate on every task. Accepted concepts supply cues to
the small model, making learned abstractions available without repeatedly
consulting the LLM. Meanwhile, complete original experiences remain in a
cold archive, while redundant mastered cases can leave the active working
set. This distinguishes \emph{internalizing} experience into operational
knowledge from merely summarizing or deleting historical records.

The main contributions of this work are summarized as follows:
\begin{itemize}
  \item \textbf{Small-model-first learning with selective LLM reasoning.}
  We integrate feedback-triggered learning and evidence-triggered historical
  reviews into an ordinary lightweight prediction pathway, so successful
  routine processing does not inherently require LLM inference.

  \item \textbf{Experience-driven, independently validated concept
  abstraction.} We allow an LLM to propose shared observable conditions
  from temporally separated historical experiences and internalize accepted
  concepts as trainable hierarchical neural representations, rather than
  treating textual proposals as established knowledge.

  \item \textbf{Long-term knowledge reuse with active--cold memory.}
  We combine learned concept cues with selective active-experience retention
  and complete archival preservation, and evaluate prediction, concept
  generalization, and memory costs in complementary simulated agent tasks.
\end{itemize}

The remainder of this paper is organized as follows.
Section~\ref{sec:related} reviews efficient model collaboration,
experience-based reflection, and concept and memory learning.
Section~\ref{sec:model} presents the SCALE mechanisms.
Section~\ref{sec:verification} describes the experimental settings and
verifies selective learning, delayed-evidence reasoning, concept
internalization, and retrospective abstraction.
The final section \ref{sec:conclusion} concludes the paper.

\section{Related Work}
\label{sec:related}

\subsection{Efficient Small--Large Model Collaboration}
\label{subsec:related_collaboration}

Large language models offer flexible reasoning but can impose substantial
inference cost when called for every observation or action. Prior work on
cost-aware inference has consequently explored the allocation of different
requests to models of different capabilities. FrugalGPT studies prompt
adaptation, model approximation, and cascaded inference, demonstrating that
selective access to more expensive language models can substantially reduce
cost while maintaining task performance~\cite{xin2025hybrid}.
RouteLLM learns a router from preference data to choose between stronger and
weaker language models~\cite{ong2025routellm}. These approaches establish an
important operational principle: the most capable model need not process
every input. Their principal decision, however, concerns \emph{which model
should answer the current query}, rather than \emph{what reusable knowledge
should be learned from an intervention}.

A complementary direction transfers knowledge from a powerful teacher to a
compact student. Knowledge distillation replaces an expensive predictor or
ensemble with a smaller network trained to reproduce informative teacher
outputs~\cite{hinton2015distilling}. DistilBERT illustrates the utility of
this principle for language representations by producing a lighter model
through a dedicated distillation procedure~\cite{sanh2019distilbert}. These
methods offer efficient deployment and make a strong case for internalizing
high-capacity knowledge into smaller parameters. Nevertheless, conventional
distillation is usually organized around a prepared teacher--student
training process; it does not, by itself, specify how an autonomous agent
should retrieve temporally separated experiences, discover a new concept,
and decide when the teacher needs to reason again.

SCALE combines these otherwise complementary perspectives without treating
them as interchangeable. A lightweight task model is the ordinary decision
path; independently observed errors and accumulated cross-episode concept
evidence are separate reasons for selective LLM intervention. Instead of
simply switching to the LLM's answer or distilling its unconstrained output,
the framework requests causal or conceptual hypotheses, subjects proposed
training examples to independent environment validation, and incorporates
accepted knowledge into lightweight task and concept networks. The distinction
is thus between \emph{cost-aware answer selection} and \emph{selective
reasoning that changes the agent's future small-model capabilities}. This
design does not assume that the LLM invocation rate must decrease
monotonically or that model selection alone guarantees reliable knowledge.

\subsection{Experience-Based Reflection and Reusable Agent Skills}
\label{subsec:related_reflection}

Experience-based LLM agents show how environmental feedback can become a
resource for subsequent reasoning. Reflexion converts trial outcomes into
verbal reflections stored in episodic memory, improving later attempts
without updating the LLM's weights~\cite{shinn2023reflexion}. ExpeL
collects experiences from training tasks, extracts natural-language insights,
and recalls both experiences and insights at inference time
~\cite{zhao2024expel}. Their main advantage is that lessons from failure can
be used with closed or otherwise non-trainable foundation models. A limitation
for resource-constrained autonomous systems is that the knowledge is primarily
consumed through retrieved text and renewed LLM inference, rather than
becoming a compact neural decision mechanism that independently predicts
routine actions.

More ambitious agents explicitly combine experience, abstraction, and
long-horizon behavior. Generative Agents maintain a record of observations,
synthesize higher-level reflections, and retrieve memories for planning,
which supports coherent behavior across extended interactions
~\cite{park2023generative}. Voyager uses environment feedback and an
evolving executable skill library to reuse acquired behavior in an open-ended
embodied environment~\cite{wang2024voyager}. These studies are particularly
relevant because they show that experience need not remain an isolated
example: it can be reorganized into reflections or reusable skills. Their
representations nevertheless serve different purposes from SCALE's neural
concept cues. A textual reflection or retrieved program can guide an LLM
agent, while a learned parent concept is intended to affect the predictions
of the lightweight model itself.

A recurring methodological issue is that a plausible reflection is not
automatically a valid rule. Summaries can omit boundary conditions, and a
generated example can conflict with the environment it purports to describe.
SCALE therefore separates LLM \emph{proposal} from \emph{acceptance}: error
explanations and positive/negative counterfactual cases are checked against
independent event semantics before they are used for training; proposed
hierarchical concepts require additional held-out validation. Historical
recall is used to compare observations separated by distractors or long
delays, and evidence can trigger a concept review even in the absence of an
immediate action error. This extends experience-based reflection toward
validated knowledge acquisition, rather than presenting reflective text
alone as evidence that the small model has learned.

\subsection{Concept Representation, Continual Learning, and Long-Term Memory}
\label{subsec:related_concepts}

Concept-based neural models introduce structured intermediate knowledge
between observations and decisions. Concept Bottleneck Models predict
human-specified concepts before predicting a task label, making decisions
more interpretable and allowing interventions on concept values
~\cite{koh2020cbm}. Concept Embedding Models learn richer concept-conditioned
representations and improve the flexibility of concept-based prediction,
including settings with incomplete concept supervision
~\cite{zarlenga2022cem}. These approaches provide useful precedents for
making concepts computationally actionable rather than merely recording
them in natural language. Their standard setting, however, assumes a
provided concept vocabulary or some concept supervision. SCALE instead
studies how an agent can \emph{propose} candidate concepts from accumulated
episodes and subsequently learn verified concept detectors and parent nodes.
The approach remains dependent on observable features and validation; an
LLM-generated concept is not accepted merely because it sounds plausible.

Long-term memory research addresses a different, closely related problem:
how an agent can retain and access relevant history. MemoryBank uses memory
updating and selective reinforcement to maintain personal conversational
context over time~\cite{zhong2024memorybank}. HippoRAG organizes knowledge
for efficient long-range retrieval with a graph-based approach
~\cite{gutierrez2024hipporag}. More recently, A-Mem dynamically links and
reorganizes structured memory notes as new experiences arrive
~\cite{xu2025amem}. These methods advance retrieval, organization, and the
evolution of external memories; in particular, A-Mem already treats memory
organization as an adaptive process. The distinction from SCALE is therefore
\emph{not} that earlier systems lack abstraction or dynamic memory. It is
that their memory structures are chiefly consumed through retrieval-based
LLM processing, whereas SCALE additionally tests whether validated
experience-level relations can become reusable lightweight predictive
parameters.

Continual-learning approaches provide another route to persistent
knowledge. Gradient Episodic Memory, for example, retains selected past
examples and constrains updates to reduce forgetting while learning new
tasks~\cite{lopezpaz2017gem}. Such methods are valuable for preserving past
competence, but replaying examples and learning a shared concept are not
equivalent operations: different object classes may support one abstract
action condition, and a collection of correct predictions need not identify
that condition explicitly. SCALE keeps a complete cold archive while
selecting a smaller active working set; abstract concepts can represent
repeated experience patterns, and the archive remains available for later
retrieval. This design does not imply that compression always improves
wall-clock lookup latency or that concept generalization is error-free.

Taken together, existing studies establish the benefits of economical
inference, reflective experience reuse, concept-guided neural prediction,
and adaptive memory. SCALE focuses on their intersection: \emph{small-model-
first operation, on-demand LLM reasoning over historical evidence,
independently validated hierarchical concept formation, and internalization
into lightweight predictors with active--cold experience management}. The
experimental question is whether these mechanisms yield useful knowledge
and reduced active-memory requirements, rather than whether every newly
introduced concept layer must immediately outperform its predecessor on
every metric.

\section{Small-Model-First Concept Abstraction and LLM-Guided Experience Internalization - The SCALE Model}
\label{sec:model}

We propose \emph{SCALE} (\textbf{S}mall-model-first \textbf{C}oncept
\textbf{A}bstraction with \textbf{L}LM-guided \textbf{E}xperience
Internalization), a continuing learning framework in which the small model
handles ordinary decisions and a large language model (LLM) is invoked
selectively to reason about failures or accumulated evidence that may reveal
reusable knowledge. Unlike an agent that repeatedly queries an LLM for each
action, SCALE seeks to transfer the useful results of expensive reasoning into
lightweight predictors. Unlike a method that stores every experience in its
active context, SCALE also uses learned concepts to retain representative
experiences actively while preserving all original records in an archive.

\begin{figure*}[t]
\centering
\resizebox{\textwidth}{!}{%
\begin{tikzpicture}[
  every node/.style={font=\footnotesize},
  component/.style={draw,rounded corners=2pt,align=center,
    text width=2.42cm,minimum height=1.08cm,inner sep=5pt},
  arrow/.style={->,>=stealth,semithick},
  feedback/.style={->,>=stealth,semithick,dashed}]
  \node[component] (obs) at (0,1.50) {Observed event\\sequence and context};
  \node[component] (small) at (3.20,1.50) {Small-model\\action prediction};
  \node[component] (outcome) at (6.40,1.50) {Action and independent\\outcome feedback};
  \node[component] (archive) at (9.60,1.50) {Experience archive\\and active working set};
  \node[component] (gate) at (12.80,1.50) {Error or accumulated\\concept evidence?};
  \node[component] (reason) at (12.80,-1.35) {Selective LLM\\reflection/proposal};
  \node[component] (verify) at (9.60,-1.35) {Counterfactual and\\concept validation};
  \node[component] (learn) at (6.40,-1.35) {Task-model and\\neural concept training};
  \node[component] (internal) at (3.20,-1.35) {Internalized concept cues\\and active-memory update};
  \draw[arrow] (obs)--(small);
  \draw[arrow] (small)--(outcome);
  \draw[arrow] (outcome)--(archive);
  \draw[arrow] (archive)--(gate);
  \draw[feedback] (gate)--node[right,font=\scriptsize] {on demand} (reason);
  \draw[arrow] (reason)--(verify);
  \draw[arrow] (verify)--(learn);
  \draw[arrow] (learn)--(internal);
  \draw[feedback] (internal)--node[right,font=\scriptsize] {reuse} (small);
\end{tikzpicture}}
\caption{Overview of SCALE. Every interaction is processed by the small
model and subsequently recorded with independent feedback. Only selected
errors or accumulated concept evidence activate the LLM. Candidate examples
and concepts are validated before their effects are internalized into the
small model; complete historical records remain available in cold storage.}
\label{fig:scale_overview}
\end{figure*}
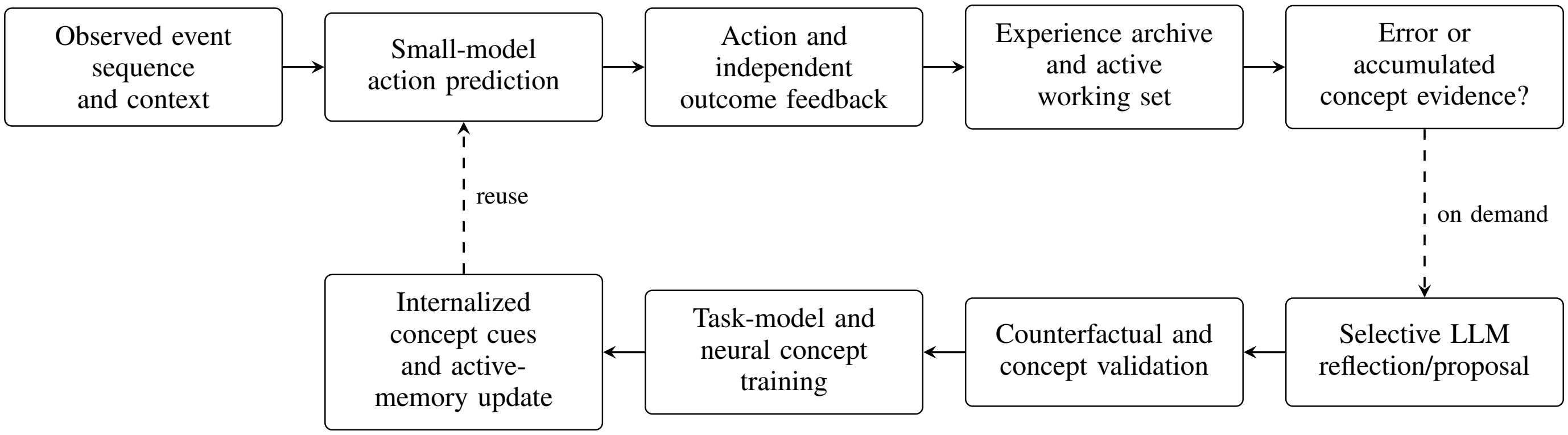

The model connects four functions: (i) small-model-first action prediction,
(ii) feedback- and evidence-triggered LLM reflection, (iii) independently
validated hierarchical concept formation, and (iv) concept internalization
and active--cold memory management. A concept may progress from a relation
among several concrete experiences to a more general relation supported by
later observations. The emergence of a higher-level concept is treated as
knowledge-structure growth, not as a guarantee that every higher-level
classifier must outperform its lower-level counterpart. Figure~\ref{fig:scale_overview}
illustrates the overall interaction.

\subsection{Small-Model-First Prediction and Experience Recording}
\label{subsec:scale_small}

At interaction step $t$, the agent receives an observable situation
$s_t=(x_{t,1:L_t},z_t)$, where $x_{t,1:L_t}$ is an ordered event sequence and
$z_t$ contains available context or sensor features. Informative events and
the eventual decision may be separated by distracting observations. The
agent must select an action from $\mathcal A$ before the correct action is
known. Its independently assessed outcome $a_t^*$ becomes available only
after the decision. We store an experience as
$e_t=(s_t,\hat a_t,a_t^*,q_t)$, where $q_t$ denotes the predictor's
confidence or score summary. Historical records may also contain a short
reasoning summary if an LLM intervention occurred; simulator-internal causes
and future evaluation labels are never used as routine predictor inputs.

The small task model $f_\theta$ yields base logits $b_t$ from the current
situation. An optional concept network $h_\phi$ produces learned cues
$c_t=h_\phi(s_t;\mathcal G_t)$ from an evolving concept graph $\mathcal G_t$.
The task-model adapter $r_\theta$ then combines both sources:
\begin{equation}
\label{eq:scale_prediction}
\begin{aligned}
 b_t &= f_\theta(s_t), \qquad p_t^0=\operatorname{softmax}(b_t),\\
 \ell_t &= b_t+r_\theta(c_t,p_t^0)-r_\theta(\mathbf 0,p_t^0),\\
 \hat a_t &= \arg\max_{a\in\mathcal A}[\ell_t]_a.
\end{aligned}
\end{equation}
The subtraction ensures that a zero cue vector leaves the base prediction
unchanged. Therefore, before any concept has been accepted, the ordinary
small-model pathway is still usable. Concept scores subsequently influence
that pathway instead of replacing action prediction with an LLM response.
The evaluated implementation uses a bidirectional GRU with temporal
attention for $f_\theta$ and a trainable cue-conditioned residual adapter;
SCALE's small-model-first principle does not require this particular
network architecture.

\subsection{On-Demand LLM Reflection and Verified Learning}
\label{subsec:scale_reflection}

SCALE distinguishes \emph{two reasons} to request LLM reasoning. First,
when independent feedback reveals a wrong prediction
($\hat a_t\neq a_t^*$), the agent may ask the LLM to identify an
observable cause and produce contrastive learning material. Second, a
review may be initiated when the archive accumulates potentially shared
positive and negative evidence across distinct situations, even if the
latest action was not wrong. The latter path allows concept formation to
arise from accumulated experience rather than requiring a task failure at
the exact moment of abstraction. Both types of intervention are subject to
finite budgets; a routine successful prediction does not inherently require
an LLM call. The current experimental implementation evaluates
feedback-driven repair and evidence-driven reviews, but does not establish a
general uncertainty-only trigger.

For error-driven reflection, the agent supplies the LLM with the observed
sequence, the small-model decision, the independent feedback, and selected
\emph{previously verified} experiences. When enabled, retrieval searches
the complete historical archive so that evidence separated by long time
intervals can still be compared. The LLM proposes an explanatory factor,
positive examples in which it is active, and negative examples that isolate
its absence or a corrective condition. These outputs are hypotheses, not
additional ground-truth labels. Let $\widetilde{\mathcal D}_t$ be the proposed
examples and $V$ an independent consistency checker. The usable training
set is restricted to
\begin{equation}
\label{eq:scale_verified_examples}
\mathcal D_t^{\mathrm{valid}}=
\{d\in\widetilde{\mathcal D}_t\mid V(d)=1\}.
\end{equation}
The checker evaluates both positive and negative examples against the
environment's event semantics and rejects contradictory or unsupported
labels. A response cannot train the model merely because the LLM returned
syntactically valid text. When accepted training examples and learned cues
are present, the task model is optimized using
\begin{equation}
\label{eq:scale_task_loss}
\mathcal L_{\mathrm{task}}=
\frac{\operatorname{CE}(\ell,a^*)+
\lambda\operatorname{CE}(b,a^*)}{1+\lambda},
\end{equation}
where $\operatorname{CE}$ denotes cross-entropy and $\lambda\geq0$ weights
the base prediction objective. The auxiliary base loss maintains a useful
non-concept decision path; ordinary supervised learning is used in the
absence of accepted concept cues. Consequently, the computationally
expensive reasoning is converted into parameter updates that can be reused
without rerunning the same LLM analysis for each subsequent observation.

\subsection{Experience-Driven Hierarchical Concept Formation}
\label{subsec:scale_concepts}

Repairing individual errors is not sufficient to create reusable knowledge:
the system must also recognize conditions shared by episodes that originally
appear unrelated. Consider a simulated agent that learns risk from contact
with sharp glass and stone fragments. Their observed co-occurrence of sharp
edges and likely contact may motivate a concept that applies across object
types. Later experience with a flexible sharp fragment can motivate a
broader hypothesis in which material rigidity is no longer required.
Smooth edges, protected contact, and avoided contact are important negative
cases. This example illustrates concept \emph{formation and extension}, not
an assumption that every generalized hypothesis is necessarily correct.

During a concept review, the LLM receives historical \emph{training-only}
examples with their observed event sequences and outcome feedback, together
with any previously accepted concept signature. It proposes a named concept,
a short mechanism explanation, references to actual historical cases,
candidate necessary and excluded properties, and possible falsification
cases. Let $O(s)$ denote the observable property representation extracted
from the situation, and let $P_C$ and $N_C$ denote respectively the required
and excluded properties of a candidate concept $C$. Its interpretable
signature is
\begin{equation}
\label{eq:scale_signature}
I_C(s)=\mathbb I[P_C\subseteq O(s)]\,
\mathbb I[N_C\cap O(s)=\varnothing].
\end{equation}
The signature is restricted to the supplied observable-feature vocabulary;
concrete object identities may identify source experiences but are not valid
concept conditions. Importantly, $I_C$ is a \emph{proposed explanatory
condition}, not a replacement for evaluating temporal order and action
outcomes. The event-sequence model remains responsible for those temporal
relationships.

A proposed signature is not immediately deployed. The evidence is separated
into historical training and independent concept-validation portions, with
positive and negative examples covering relevant categories and boundary
cases. Candidate concepts are screened using discrimination criteria such
as F1 and false-positive rate, sufficient distinct evidence patterns, and,
for adaptive evidence reviews, conservative bounds on class-conditional
reliability. A neural parent detector must also pass held-out validation.
Unsuccessful proposals remain unaccepted; final hold-out and transfer cases
are excluded from concept proposal, training, and acceptance. Thus the LLM
\emph{proposes} a generalization, while independent evidence and a learned
small-model representation determine whether it becomes operational.

An accepted concept is added to a graph $\mathcal G_t$ containing neural
leaf detectors and trainable parent detectors. Let $u_i(s)$ be a learned
leaf activation for factor $i$, and let $v_j(s)$ be a parent activation. For
a leaf with immediate parent $j(i)$, the hierarchy can provide the gated
activation
\begin{equation}
\label{eq:scale_hierarchical_cue}
\widetilde u_i(s)=u_i(s)\,v_{j(i)}(s)
\prod_{k\in\operatorname{Anc}(j(i))}v_k(s),
\end{equation}
where $\operatorname{Anc}(j(i))$ denotes accepted higher ancestors. Parent
thresholds also permit conditional evaluation of child detectors, and
parent-level action cues can be supplied directly to the task adapter in
Eq.~\eqref{eq:scale_prediction}. A newly learned parent need not erase an
older parent: a later accepted higher-level concept can coexist with the
earlier representation and provide another route to reusable action
knowledge. In this sense, hierarchical abstraction is more than storing a
list of related episodes, because the parent itself is trained on verified
contrastive examples and is available at inference time.

\subsection{Knowledge Internalization and Active--Cold Memory}
\label{subsec:scale_memory}

SCALE internalizes a validated concept in two complementary forms. Its
interpretable signature and historical evidence references remain
inspectable, while trainable concept detectors and the small-model adapter
learn how the concept influences future predictions. This distinction is
important: merely receiving a plausible LLM explanation does not mean that
the small model has learned it. The learned cue must survive independent
validation and be useful as part of subsequent lightweight inference.

For long-term operation, the model maintains a complete archive
$\mathcal M_t^{\mathrm{archive}}$ and an active experience set
$\mathcal M_t^{\mathrm{active}}\subseteq
\mathcal M_t^{\mathrm{archive}}$.
Every completed interaction is preserved, including incorrect decisions and
counterexamples. Repeated confidently mastered cases may be retired from
the active set; after a concept has been validated, additional redundant
cases matching the learned conditions can also be made inactive. However,
representative positive examples, informative errors, and negative boundary
cases are retained to avoid making the working set consist only of easy
positives. Retired experiences are \emph{not deleted}: retrieval can access
the complete archive and temporarily reactivate relevant cold records when
future learning or reasoning needs them.

The representation used for memory abstraction need not be identical in
every experiment. A feedback-mined, single-tier concept predicate can supply
useful cues and support active-memory reduction, whereas the integrated
SCALE configuration additionally evaluates LLM-proposed signatures and
neural parent concepts. This distinction prevents memory compression from
being attributed to an LLM-produced hierarchy where that hierarchy was not
actually used. Memory efficiency is characterized by the active-record
count, effective active storage, and retrieval-candidate count; these are
reported separately from measured wall-clock retrieval latency, because
reducing active storage does not by itself guarantee faster search.

Overall, SCALE establishes an experience-to-knowledge path: the small model
acts first; feedback enriches the permanent historical archive; selected
errors or cross-experience evidence call for deeper LLM reasoning;
independently validated examples and concepts are learned by small models;
and mastered experiences can move from active memory to cold storage.
Section~\ref{sec:verification} evaluates these elements through selective
error learning, delayed-evidence attention, concept-based memory reduction,
and LLM-guided hierarchical concept formation. The goal is autonomous
growth of reusable small-model knowledge, rather than continuous reliance
on LLM predictions or the requirement that each newly formed abstraction
immediately improve every evaluation metric.

\section{Verification}
\label{sec:verification}

\subsection{Experimental Environment and Evaluation Protocol}
\label{subsec:verification_settings}

\textbf{Objectives and simulated scenarios.}
We evaluate a small-model-first autonomous learning system in which a compact
predictor acts on routine observations, while an LLM is selectively invoked
to investigate errors and synthesize experience-based abstractions. The goal
is not to make every newly formed higher-level concept immediately more
accurate than its lower-level components. Instead, we ask whether the system
can learn from feedback, recall temporally separated evidence, form and
validate reusable concepts, internalize them into neural predictors, and
manage accumulated experiences more economically.

The environment is a \emph{symbolic, event-based robot-agent simulation},
not a physical-robot or raw-image recognition experiment. Every example
contains ordered observations, an action decision, and independent outcome
feedback. The simulator covers (i) contact safety, (ii) low-friction grasp
adjustment, and (iii) uncertain visual observation. In the safety scenario,
a sharp glass or stone fragment may require \texttt{AVOID\_SHARP\_CONTACT}
when contact is likely, whereas smooth edges, avoided contact, or protective
gear are negative/boundary cases. Ceramic and metal fragments are reserved
for transfer. In the grasp scenario, a wet cup or polished box with low
friction and a planned grasp may require \texttt{ADJUST\_GRIP}; high-friction
objects and unplanned grasps provide negatives, while oily bottles and
smooth cans are unseen subtypes. In the perception scenario, glare and
shadows can justify \texttt{REOBSERVE} when a visual decision is necessary;
adequate contrast or a nonvisual task does not. Fogged and backlit scenes
serve as unseen subtypes. Some observations and their consequences are
separated by distracting events, so the correct response need not be
recoverable from the final event alone. These scenarios measure whether
concepts capture combinations of properties rather than merely object names.
The environment oracle produces independent feedback and validates proposed
counterfactuals, but is never used as a predictor input.

\textbf{Methods and configurations.}
The default task predictor is a bidirectional gated recurrent neural network
(GRU) with a trainable concept-cue adapter. Candidate attention and parent
concept models are neural, with a Transformer-style attention mechanism.
A live DeepSeek-compatible LLM is invoked when the configured error or
concept-evidence policy requests additional thinking; it is not used for
every routine prediction. A candidate LLM proposal specifies observable
necessary/absent conditions, historical evidence references, and possible
positive/negative tests. Generated learning materials must agree with an
independent environment validator, and a proposed parent becomes deployable
only after held-out neural validation.

The Formal evaluation comprises 10 seeds, using 320 initial
training examples, 96 validation examples, 360 online
events, and 180 hold-out instances in the task experiments.
Exp.~2 uses observation--decision gaps of 4, 32, and
96 intervening events. Exp.~3 uses 6000 historical
experiences, with the initial 0.55 fraction devoted to known
subtypes; later online blocks vary new-subtype proportions. Exp.~4 uses
gap 96, introducing a flexible sharp polymer fragment only after
initial glass/stone evidence; independent ceramic/metal and difficult
negative combinations are not used to train or accept the concept. A
separate final concept probe contains 150 challenging samples
per seed, with standard cases excluded. Memory timing is evaluated on
4000 episodes using identical frozen weights and observations.

The \texttt{task\_matched} setting equalizes the task-model update
protocol but allows extra neural attention/concept optimizer steps and
LLM calls. We therefore report optimizer steps, logical requests,
actual HTTP attempts, and storage/timing separately. All reported averages
use independent seeds as the unit of replication. Accuracy is the fraction
of correct action decisions; transfer accuracy measures independently
constructed unseen subclasses and condition combinations. Macro-F1 measures
class-balanced performance where relevant. Concept depth and verified
concept count quantify representation growth; active memory includes
routinely accessible history, whereas the full cold archive remains
available. Effective active bytes account for record/index data plus
shared learned weights, not process RSS. Retrieval P50/P95 denote measured
latency percentiles. The paired confidence intervals are Student-$t$
intervals on the ten seed-level differences at 95\% confidence.

\subsection{Exp.~1: Selective Reasoning and Experience Reuse}
\label{subsec:verification_e1}

The initial hypothesis is that errors can drive selective LLM reasoning and
small-model retraining, without forwarding every ordinary input to an LLM.
The simulation presents a sequence of object/contact, grasp, and visual
conditions. The small model predicts first and receives independent
feedback; incorrect outcomes can trigger reasoning and verified
counterfactual material. \texttt{FixedSmallModel} is the unadapted
predictor; \texttt{ContinualReplay} performs ordinary experience replay
without LLM; \texttt{OursRepairOnly} reasons about the current error;
\texttt{OursRepairWithRecall} additionally consults related verified
history; and \texttt{OursFlatAttention} learns flat condition attention.

\begin{table}[t]
\centering
\caption{Exp.~1: independent action accuracy and logical LLM calls,
means over ten seeds.}
\label{tab:e1}
\resizebox{\columnwidth}{!}{%
\begin{tabular}{lrrr}
\toprule
Method & Hold-out (\%) & Transfer (\%) & LLM calls\\
\midrule
FixedSmallModel & 87.11 & 84.67 & 0.0\\
ContinualReplay & 95.39 & 96.28 & 0.0\\
OursRepairOnly & 97.39 & 96.28 & 14.8\\
OursRepairWithRecall & \textbf{98.28} & 96.50 & 13.0\\
OursFlatAttention & 97.17 & \textbf{96.89} & 13.5\\
\bottomrule
\end{tabular}}
\end{table}

Table~\ref{tab:e1} shows that recalling past experiences yields
98.28\% hold-out accuracy versus 95.39\%
with continual replay and 87.11\% with the fixed model.
The 2.89-percentage-point gain over replay has
a paired interval entirely above zero (Table~\ref{tab:paired}).
Recall improves current-error-only repair from 97.39\%
to 98.28\%, but the smaller paired difference has an
interval crossing zero. These results support selective LLM-assisted
learning and show a benefit of retained experience in this setting;
they do not prove that the intervention rate decreases autonomously as
knowledge grows, since the policy also imposes an invocation budget.

\subsection{Exp.~2: Learning with Long Delays and Distractors}
\label{subsec:verification_e2}

Here we ask whether the predictor can associate an early observation with
a later consequence. We construct otherwise comparable event sequences
with gaps of 4, 32, and 96; irrelevant events intervene
between the informative condition and the decision. A fixed small model,
flat learned attention, learned hierarchical attention, and a random
hierarchy are compared using independent hold-out episodes. The random
hierarchy tests whether merely organizing detectors in a tree is enough,
rather than learning meaningful parent structure.

\begin{table}[t]
\centering
\caption{Exp.~2: hold-out action accuracy (\%) at three delayed-evidence gaps.}
\label{tab:e2}
\resizebox{\columnwidth}{!}{%
\begin{tabular}{lrrr}
\toprule
Method & Gap 4 & Gap 32 & Gap 96\\
\midrule
FixedSmallModel & 87.11 & 52.17 & 36.17\\
FlatAttention & 89.83 & \textbf{89.33} & 86.39\\
HierarchicalAttention & \textbf{90.61} & 86.94 & 87.00\\
RandomHierarchy & 89.83 & 86.61 & \textbf{87.44}\\
\bottomrule
\end{tabular}}
\end{table}

\begin{figure}[t]
\centering
\includegraphics[width=\columnwidth]{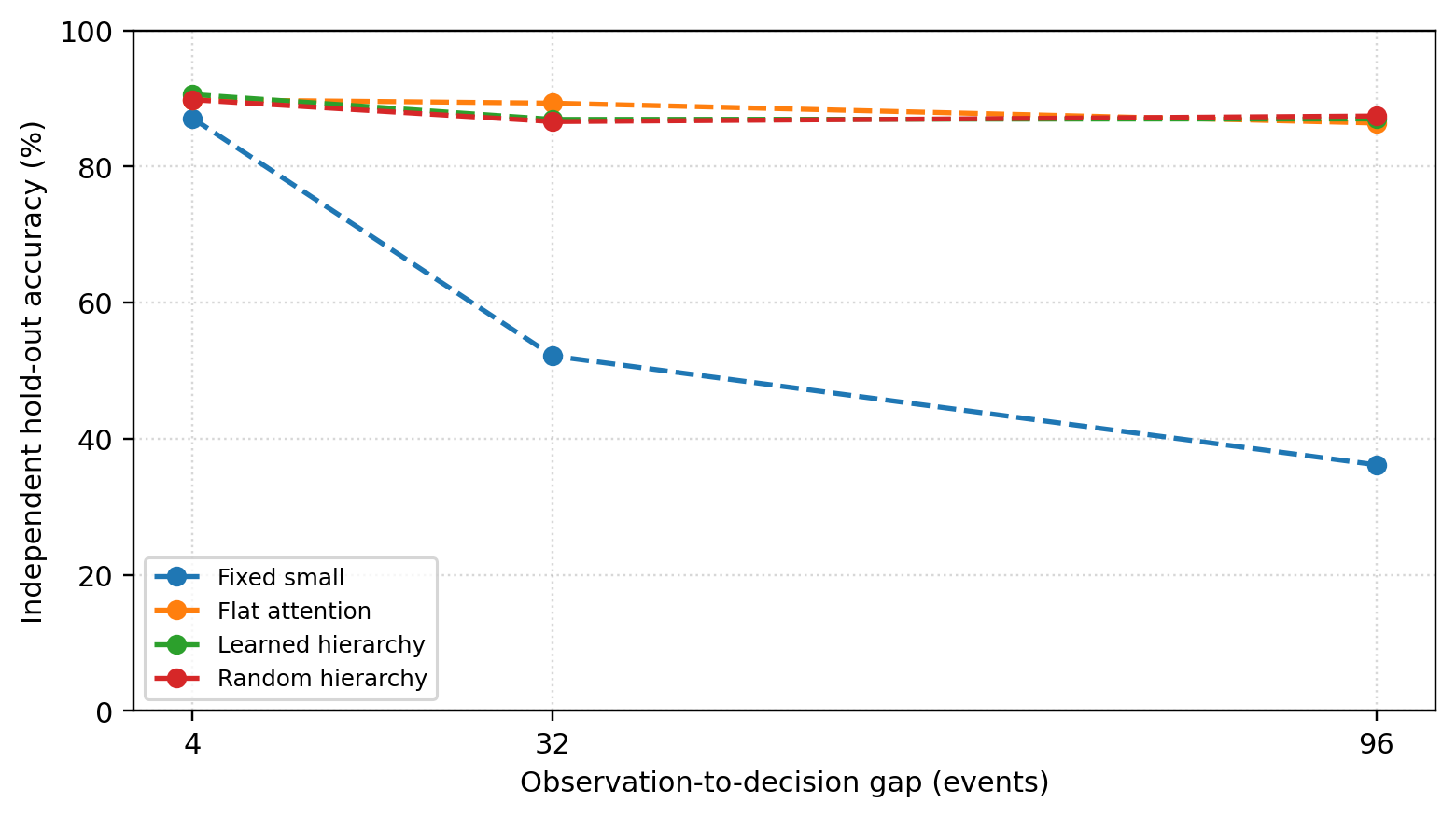}
\caption{Exp.~2 delay sensitivity. All line plots use dashed strokes;
points reproduce Table~\ref{tab:e2}.}
\label{fig:e2}
\end{figure}

As Table~\ref{tab:e2} and Fig.~\ref{fig:e2} show, fixed-model accuracy
falls from 87.11\% at gap 4 to 36.17\%
at gap 96, whereas learned hierarchical attention retains
87.00\%. This supports learning from delayed evidence.
However, the learned hierarchy reaches 86.94\% at gap
32, below the flat method's 89.33\%, and the random
hierarchy marginally exceeds the learned hierarchy at gap 96
(87.44\% against 87.00\%). Hierarchical structure
is therefore feasible, but its independent accuracy advantage over
flat or random organization is not universal.

\subsection{Exp.~3: Internalized Concepts and Active-Memory Reduction}
\label{subsec:verification_e3}

When similar decisions repeat over a long lifetime, an agent should be able
to use the learned common-condition relation without keeping every training
example in active memory. After known-subclass warm-up, the simulator
introduces new-subclass fractions of 0\%, 25\%,
50\%, and 75\% in successive online blocks. The
first frozen transfer probe precedes all feedback on the new categories;
later blocks permit online adaptation and are not zero-shot tests.
Protected contact, smooth edges, high friction, and unnecessary
reobservation make negative decisions nontrivial. We compare
\texttt{FullHistory} (all records active), \texttt{OursArchivedRecall}
(cold archival and selective recall without concept compression), and
\texttt{OursAbstractArchive} (train common-condition cues, internalize
them, remove redundant mastered items from active storage, and retain
the original archive). An active-only policy and archive-without-recall
are additional information-retention controls, but not all are needed
in the representative table.

\begin{table}[t]
\centering
\caption{Exp.~3: independent task performance and active-memory footprint.
MB is decimal MB; original archives remain complete for these methods.}
\label{tab:e3}
\resizebox{\columnwidth}{!}{%
\begin{tabular}{lrrrr}
\toprule
Method & Active records & Archive & Active MB & Hold-out (\%)\\
\midrule
FullHistory & 6000.0 & 6000 & 8.965 & 91.89\\
OursArchivedRecall & 2474.9 & 6000 & 5.044 & 93.72\\
OursAbstractArchive & \textbf{1527.8} & 6000 & \textbf{3.975} & \textbf{94.83}\\
\bottomrule
\end{tabular}}
\end{table}

\begin{figure}[t]
\centering
\includegraphics[width=\columnwidth]{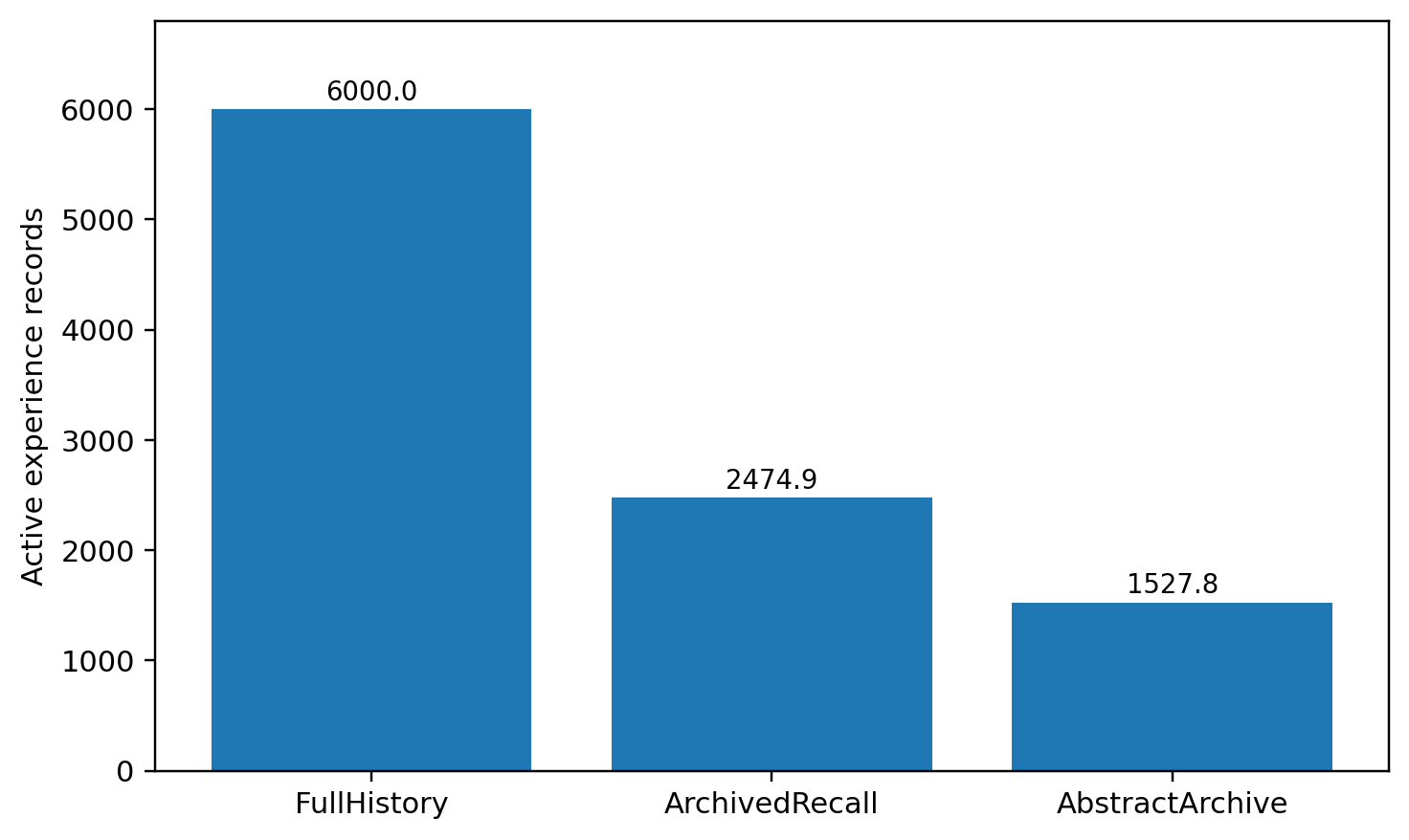}
\caption{Exp.~3 active episode counts computed from the ten raw per-seed
records. Cold archived episodes are retained.}
\label{fig:e3}
\end{figure}

Table~\ref{tab:e3} and Fig.~\ref{fig:e3} show that abstraction keeps
1527.8 routinely active records instead of
6000.0 (a 74.5\% reduction), with
3.975 rather than 8.965 MB of effective
active memory. Independent hold-out accuracy increases from
91.89\% to 94.83\%, an improvement of
2.94 percentage points, while all
6000 original experiences remain archived.
Relative to archived recall, the active subset is
38.3\% smaller. The very high frozen novel-subclass
transfer values (100.00\% for the abstract method
versus 99.94\% for full history) indicate that this
particular benchmark is near saturation.

\begin{table}[t]
\centering
\caption{Exp.~3: separate challenging frozen-model concept probe,
showing both useful cues and difficult negative boundaries.}
\label{tab:e3hard}
\resizebox{\columnwidth}{!}{%
\begin{tabular}{lrr}
\toprule
Cue mode & Challenging accuracy (\%) & Hazard FPR (\%)\\
\midrule
No concept cue & 35.33 & 60.43\\
Internalized concept cue & 42.53 & 70.97\\
\bottomrule
\end{tabular}}
\end{table}

The more difficult probe (Table~\ref{tab:e3hard}) raises the measured
challenging accuracy from 35.33\% without a learned concept cue
to 42.53\% with one, but its hazard false-positive rate also
rises from 60.43\% to 70.97\%.
Thus learning generalized knowledge can broaden coverage while leaving
imperfect boundaries. This is reported as a limitation, without making
future concept self-correction an additional experimental contribution.

\subsection{Exp.~4: Retrospective LLM Concept Formation}
\label{subsec:verification_e4}

Exp.~4 asks whether the LLM can propose a genuinely revised common
observable condition by reconsidering several past experiences,
rather than the program simply counting shared attributes. Sharp glass
and stone cases are encountered first; a later flexible sharp polymer
subclass supplies evidence that material rigidity may not be a necessary
condition of risk. Ceramic, metal, and difficult condition combinations
remain independent test cases. The trained model is also evaluated on
long-gap, object-switch, state-reversal, sensor-conflict, partial-evidence,
and contact-boundary negative episodes. Only proposed concepts that pass
independent positive/negative and held-out validation become neural
parent nodes. Suggested falsification descriptions are retained as
hypotheses, not represented as executed physical experiments.

\texttt{OursFull} combines small-model-first error repair, historical
recall, adaptive evidence review, LLM-proposed semantic concept signatures,
independent validation, and neural internalization. With the same broad
evidence mechanism, \texttt{OursRuleMiningAdaptive} proposes signatures
through programmed attribute mining rather than LLM reasoning.
\texttt{OursFixedEvidence} uses a fixed evidence schedule;
\texttt{OursNoRecall} omits historical recall, and
\texttt{OursFullHistoryContext} retains a much larger active history.
\texttt{OursHierarchicalAttention} offers neural hierarchy without the
complete memory mechanism; \texttt{OursRepairWithRecall} uses error-driven
reasoning and history but no abstract neural parent induction.

\begin{table*}[t]
\centering
\caption{Exp.~4 independent action results: ten-seed means. Additional
optimizer steps for concept/attention training are included.}
\label{tab:e4}
\resizebox{\textwidth}{!}{%
\begin{tabular}{lrrrrr}
\toprule
Method & Hold-out (\%) & Novel transfer (\%) & Verified parents & Mean max. depth & Optimizer steps\\
\midrule
OursFull & 89.22 & 99.00 & 2.2 & 1.8 & 1859.2\\
RuleMiningAdaptive & \textbf{91.11} & 98.44 & 2.9 & 1.7 & 1873.6\\
FixedEvidence & 86.00 & 97.56 & \textbf{3.6} & 1.9 & 1931.2\\
HierarchicalAttention & 90.67 & 96.56 & 1.8 & 1.7 & 1644.0\\
NoRecall & 90.78 & 96.94 & 0.4 & 0.4 & 1597.6\\
FullHistoryContext & 90.17 & \textbf{99.06} & 1.0 & 0.9 & 1608.0\\
RepairWithRecall & 88.11 & 98.17 & -- & -- & 800.0\\
ContinualReplay & 84.22 & 98.17 & -- & -- & 1107.2\\
\bottomrule
\end{tabular}}
\end{table*}

\textbf{A traceable LLM abstraction case.}
In seed 9, the verified hazard concept at episode
119 contains required properties
\texttt{RIGID\_FRAGMENT}, \texttt{SHARP\_EDGE}, and
\texttt{CONTACT\_LIKELY}. A later verified proposal at episode
224 removes the rigid-fragment requirement while retaining
sharpness and contact likelihood; it excludes protective gear,
contact avoidance, and smooth edges. The original recorded proposals
explicitly link the latter signature to the earlier one. Thus a live
LLM proposed a semantically more general condition and the learned
parent passed the existing independent validation protocol. This is
evidence of hypothesis formation in the predefined observable feature
space, not discovery of new latent physical properties from raw pixels.

Table~\ref{tab:e4} reports 2.2 verified OursFull
concepts per run (total 22), with second-level
abstraction reached by 8 of the ten seeds. OursFull
obtains 89.22\% routine hold-out accuracy versus
91.11\% for rule mining, a
-1.89-point difference (Table~\ref{tab:paired}).
The sign of this difference is negative: forming LLM-generated
abstractions does not automatically imply higher routine classification
accuracy. The new-category transfer accuracies
(99.00\% and 98.44\%, respectively)
are near saturation and should not replace independent difficult probes.

\textbf{Testing use of internalized concepts.}
With task weights frozen, we switch between disabled concept cues,
first-level cues, and the full learned hierarchy on the same
150 difficult nonstandard episodes per seed. This is a
within-model intervention; the rule-mining and other method comparisons
use their separately trained models but the same evaluation construction.

\begin{table}[t]
\centering
\caption{Exp.~4 independent challenging-probe accuracy. The first three
rows share identical frozen OursFull weights; other rows are different
trained methods.}
\label{tab:e4hard}
\resizebox{\columnwidth}{!}{%
\begin{tabular}{lr}
\toprule
Method / inference setting & Accuracy (\%)\\
\midrule
OursFull, no concept cues & 55.33\\
OursFull, level-1 only & 64.53\\
OursFull, full hierarchy & \textbf{66.47}\\
\midrule
RuleMiningAdaptive, full hierarchy & 62.27\\
FixedEvidence, full hierarchy & 63.40\\
HierarchicalAttention, full hierarchy & 60.47\\
\bottomrule
\end{tabular}}
\end{table}

\begin{figure}[t]
\centering
\includegraphics[width=\columnwidth]{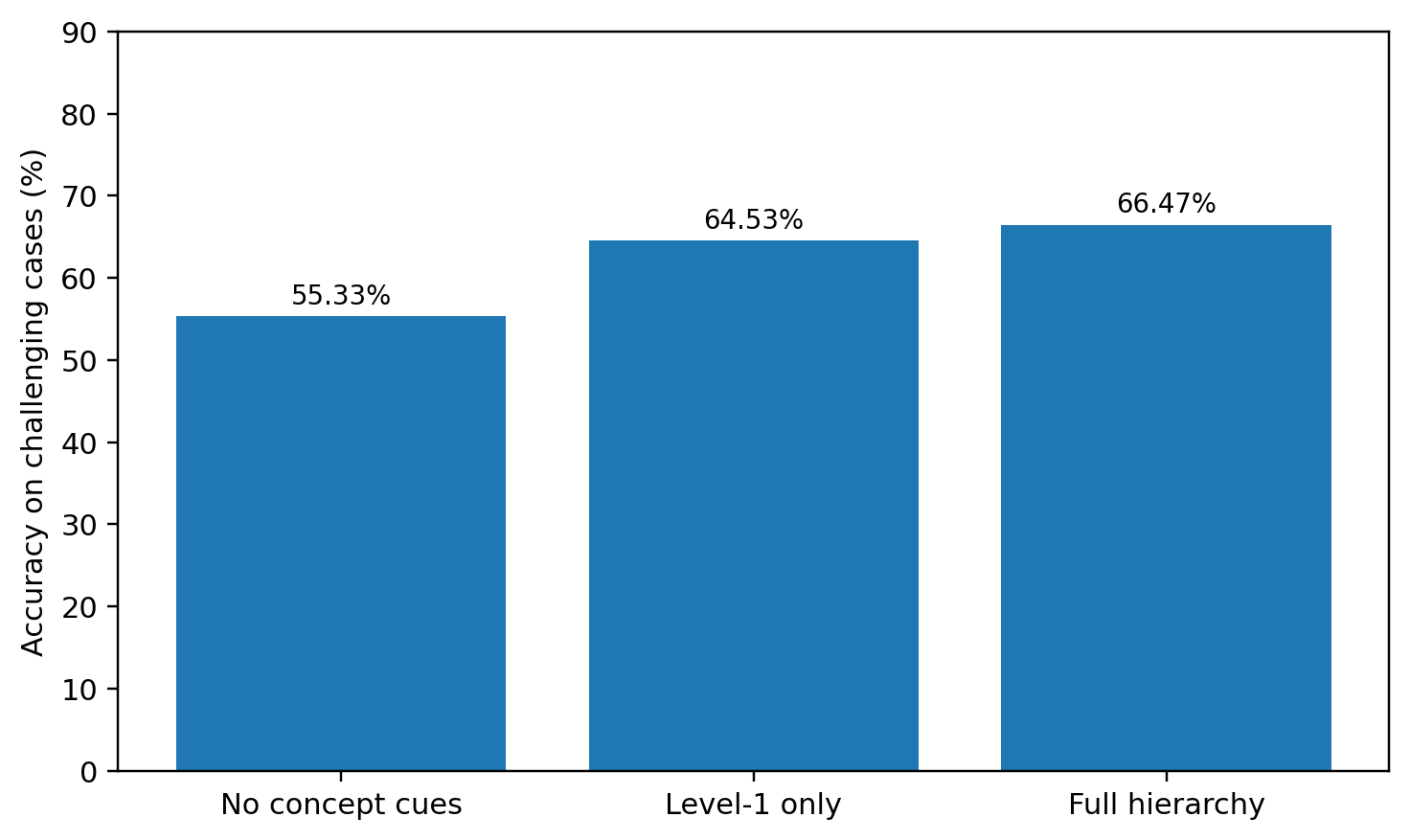}
\caption{Exp.~4: frozen OursFull task-model accuracy on challenging
cases, comparing no concept cues, first-level cues, and full learned
hierarchy (first three rows of Table~\ref{tab:e4hard}).}
\label{fig:e4}
\end{figure}

As Table~\ref{tab:e4hard} and Fig.~\ref{fig:e4} show, internalized
concept cues increase challenging accuracy from
55.33\% without cues to 66.47\% with
the full hierarchy, a gain of 11.13 points. The
1.93-point gain from first to full hierarchy is
small and its paired confidence interval includes zero.
OursFull reaches 66.47\% on the same challenging subset,
versus 62.27\% for rule mining, but its
4.20-point difference is not stable across seeds.
These outcomes demonstrate that learned abstract cues participate in
prediction while avoiding the stronger, unsupported claim that LLM
abstraction wins every accuracy comparison.

\subsection{Resource Costs, Statistical Comparisons, and Scope}
\label{subsec:verification_resources}

\textbf{Matched frozen-memory benchmark.}
We replay 4000 historical events under identical input
sequences and identical frozen neural weights. The control and full
method differ in active-memory management, not in the learned predictor.
All original episodes remain in the archive, and retrieval latencies are
measured rather than inferred from the number of candidates examined.

\begin{table}[t]
\centering
\caption{Exp.~4 paired active-memory evaluation, ten seeds. Equal frozen
accuracy is an experimental control, not a longitudinal learning result.}
\label{tab:pairedmem}
\resizebox{\columnwidth}{!}{%
\begin{tabular}{lrr}
\toprule
Metric & Hierarchy control & OursFull\\
\midrule
Active records & 3813.8 & \textbf{3459.9}\\
Effective active MB & 9.143 & \textbf{8.658}\\
Examined active records & 1961.8 & \textbf{1777.3}\\
Retrieval P50 (ms) & \textbf{1.0668} & 1.1192\\
Retrieval P95 (ms) & 2.1168 & \textbf{2.0753}\\
Frozen accuracy (\%) & 86.34 & 86.34\\
\bottomrule
\end{tabular}}
\end{table}

Table~\ref{tab:pairedmem} shows that OursFull decreases the active
subset by 9.3\%, effective active memory by
5.3\%, and examined candidates by
9.4\%. The median measured retrieval time, however,
rises from 1.0668 to 1.1192 ms (a
4.9\% increase), and the P95 decreases by only
2.0\%. Hence the evidence supports smaller active
memory and candidate lists, not a general retrieval-latency speedup.

\textbf{LLM requests and statistical uncertainty.}
OursFull issues 145 logical task-error diagnosis requests
and 111 logical history/concept requests across all
completed runs, for 256 logical requests and
336 HTTP attempts including validation retries. This
corresponds to 25.6 logical requests per
360-event run, or 7.11 requests per
100 task events when normalized by the total event count; this is
not a percentage of unique events that triggered LLM calls. The number
of requests is affected by configured budgets and past schema retries.

\begin{table*}[t]
\centering
\caption{Ten-seed paired mean accuracy differences and 95\% Student-$t$
confidence intervals, computed from original raw per-seed measurements.
Units are percentage points.}
\label{tab:paired}
\resizebox{\textwidth}{!}{%
\begin{tabular}{lrrr}
\toprule
First minus second & Mean difference & Lower CI & Upper CI\\
\midrule
Exp.~1: RepairWithRecall $-$ ContinualReplay & 2.89 & 1.92 & 3.86\\
Exp.~1: RepairWithRecall $-$ RepairOnly & 0.89 & -0.86 & 2.64\\
Exp.~3: AbstractArchive $-$ FullHistory & 2.94 & 1.59 & 4.30\\
Exp.~4: OursFull $-$ RuleMiningAdaptive (hold-out) & -1.89 & -3.64 & -0.14\\
Exp.~4: OursFull $-$ RuleMiningAdaptive (challenging) & 4.20 & -5.86 & 14.26\\
Exp.~4: OursFull full hierarchy $-$ level 1 (same model) & 1.93 & -1.07 & 4.94\\
\bottomrule
\end{tabular}}
\end{table*}

\textbf{Evidence limits.}
The four experiments substantiate selective reasoning, measurable
small-model learning, memory abstraction, and verified hierarchical
concept formation, but do not demonstrate open-ended concept invention
from raw imagery or physical-robot interaction. The task-matched
budget is not equal total computation, and the trained learned
hierarchy does not outperform flat or random controls consistently.
The archived experiences are not destroyed; memory reduction is the
size of the active working subset. Some concept generalizations
produce false positives. Finally, an LLM-output schema compatibility
correction was applied during checkpoint recovery; the completed
cohort is usable for mechanism analysis, but a strictly uniform
LLM-prompt protocol is advisable when interpreting fine-grained
request-cost differences in subsequent reproductions.

\section{Conclusion}
\label{sec:conclusion}

This paper presented SCALE, a small-model-first framework that selectively
uses large language model (LLM) reasoning to transform accumulated experiences
into reusable lightweight knowledge. Routine decisions are handled by a small
predictor, while verified errors and evidence distributed across historical
episodes trigger reflection, counterfactual example generation, and concept
proposals. Independently validated concepts are internalized as hierarchical
neural representations, and redundant mastered experiences are removed from
active memory without discarding their original records. Across ten random
seeds, historical recall increased held-out accuracy from 95.39\% for
continual replay to 98.28\%, while learned concept cues improved accuracy
on challenging cases from 55.33\% to 66.47\%. Abstract archiving reduced
active experience records from 6,000 to 1,528. These results support
experience-driven knowledge formation and more compact active memory, although
LLM-guided abstraction did not consistently outperform rule-based concept
mining and smaller active memory did not guarantee faster retrieval.


\ifCLASSOPTIONcaptionsoff
  \newpage
\fi

\bibliographystyle{IEEEtran}
\bibliography{ref}

%

\begin{IEEEbiography}{Hong Su}
  received the MS and PhD degrees, in 2006 and 2022, respectively, from Sichuan University, Chengdu, China. He is currently a researcher of Chengdu University of Information Technology Chengdu, China. His research interests include LLM and autonomous robots.
\end{IEEEbiography}




\end{document}